\documentclass[11pt]{article}

\usepackage[final]{acl}

\usepackage{times}
\usepackage{latexsym}
\usepackage{graphicx}
\usepackage{booktabs} 
\usepackage[table]{xcolor} 
\usepackage{tabularx}
\usepackage{float}
\usepackage{multirow}
\usepackage{adjustbox}
\usepackage{amsmath}
\usepackage{arydshln}
\usepackage{tikz}
\usepackage{algorithm}
\usepackage{algpseudocode}
\usepackage{pgfplots}
\usepackage[most]{tcolorbox}
\usepackage{listings}
\usepackage[all]{hypcap}
\usetikzlibrary{shapes.geometric, arrows.meta, positioning, calc, fit, backgrounds}

\usepackage[T1]{fontenc}

\usepackage[utf8]{inputenc}

\usepackage{microtype}

\usepackage{inconsolata}
\usepackage{hyperref}

\title{AXE: Low‑Cost Cross‑Domain Web Structured Information Extraction}

\author{
\begin{tabular}{c}
    \textbf{Abdelrahman Mansour\textsuperscript{1,2}} \\
    \small{\href{mailto:abdelrahman.f.mansour@gmail.com}{abdelrahman.f.mansour@gmail.com}}
\end{tabular}
\hskip 1em
\begin{tabular}{c}
    \textbf{Khaled W. Alshaer\textsuperscript{1,3}} \\
    \small{\href{mailto:khaled.w.alshaer@gmail.com}{khaled.w.alshaer@gmail.com}}
\end{tabular}
\hskip 1em
\begin{tabular}{c}
    \textbf{Motaz El-Saban\textsuperscript{1,4}} \\
    \small{\href{mailto:motaz.elsaban@gmail.com}{motaz.elsaban@gmail.com}}
\end{tabular}
\\[2ex]
\textsuperscript{1}Faculty of Computers \& Artificial Intelligence - Cairo University \\
\textsuperscript{2}Ejada Systems, \textsuperscript{3}Fawry Integrated Systems, \textsuperscript{4}Microsoft
}

\begin{document}
\maketitle
\begin{abstract}
Extracting structured data from the web is often a trade-off between the brittle nature of manual heuristics and the prohibitive cost of Large Language Models. We introduce AXE (Adaptive X-Path Extractor), a pipeline that rethinks this process by treating the HTML DOM as a tree that needs pruning rather than just a wall of text to be read. AXE uses a specialized "pruning" mechanism to strip away boilerplate and irrelevant nodes, leaving behind a distilled, high-density context that allows a tiny 0.6B LLM to generate precise, structured outputs. To keep the model honest, we implement Grounded XPath Resolution (GXR), ensuring every extraction is physically traceable to a source node. Despite its low footprint, AXE achieves state-of-the-art zero-shot performance, outperforming several much larger, fully-trained alternatives with an F1 score of 88.1\% on the SWDE dataset. By releasing our specialized adaptors, we aim to provide a practical, cost-effective path for large-scale web information extraction. Our code and adaptors are publicly available at \url{https://github.com/abdo-Mansour/axetract}.
\end{abstract}

\section{Introduction}
The World Wide Web is exponentially growing, but information is usually embedded within free-form HTML, written in different layouts with different technologies for different use cases. This makes it hard for automated data mining techniques to extract an arbitrary piece of information from an arbitrary HTML page using an arbitrary query.

\begin{figure}[t]
  \centering
  \includegraphics[width=\columnwidth]{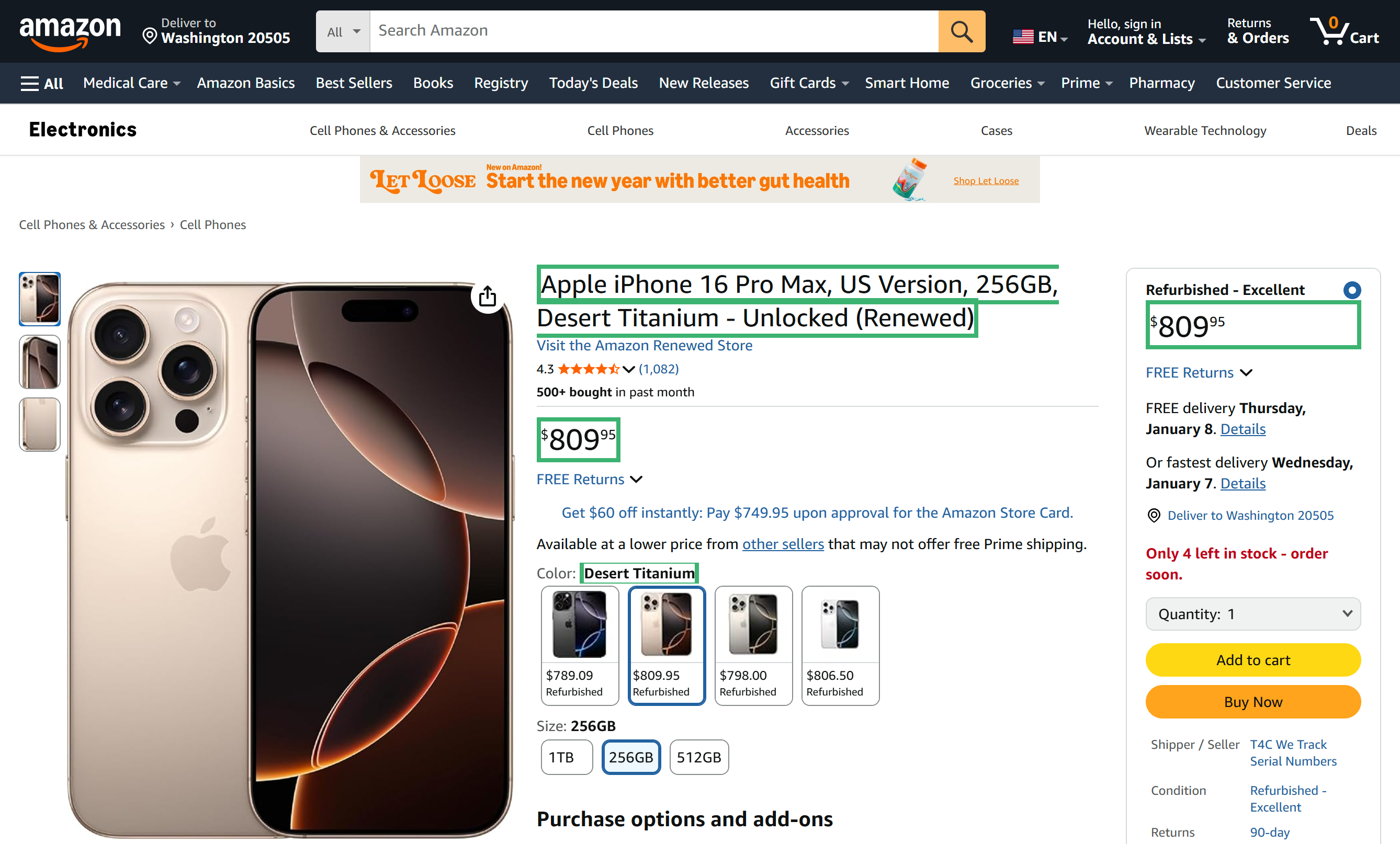}
  \caption{Example of input HTML page, the input is the raw HTML code, not the screenshot. Information requested in input schema example in Figure~\ref{fig:input_example} is bordered with green boxes}
  \label{fig:amazon_example}
\end{figure}

Our system requires 2 inputs, an HTML page and a JSON schema querying the HTML page for information. It outputs the same input JSON schema but filled with values of corresponding attributes from the page (i.e. the answers of the queries). An example of the system input can be found in Figures~\ref{fig:amazon_example}, ~\ref{fig:input_example}, while an example of the system output can be found in Figure~\ref{fig:output_example} 

An ideal information extraction solution must output a structured format, pre-defined by the user, containing the requested information with high precision and recall, and critically, it must deliver decent performance on web pages and layouts it has never seen before in the training data. For this ideal solution to be practical, it also has to be cost-effective to allow for mass data mining or agentic workflows' tools (or MCPs).

Earlier methods used heuristics and manually written CSS selectors which required a ton of manual effort and used to be custom-made for each page layout in each website individually. When deep learning models emerged, methods relied on different kinds of classifying techniques or encoding the page then searching with a query. Recently, Large Language Models (LLMs) \cite{attention_is_all_you_need, LLMs} were utilized for this task with unforeseen results. However, LLMs come at different sizes: small models' performance isn't reliable since they perform badly with big context, while large LLMs are not cost-effective for mass data mining.

\lstdefinelanguage{json}{
    basicstyle=\small\ttfamily,
    stringstyle=\color{blue},
    showstringspaces=false,
    breaklines=true,
}

\newtcolorbox{codebox}[1]{
}

\begin{figure}[h]
    \centering
    \begin{codebox}{Input Schema}
    \begin{lstlisting}[language=json]
{
  "Product Title": "",
  "Price": "",
  "Manufacturer": "",
  "Color": ""
}
    \end{lstlisting}
    \end{codebox}
    \caption{Example of an input schema, requesting information from the page.}
    \label{fig:input_example}
\end{figure}


\begin{figure}[h]
    \centering
    \begin{codebox}{Input Schema}
    \begin{lstlisting}[language=json]
{
  "Product Title": "Apple iPhone 16 Pro Max",
  "Price": "$1,039.99",
  "Manufacturer": "Apple",
  "Color": "Desert Titanium"
}
    \end{lstlisting}
    \end{codebox}
    \caption{Example of a system output, with correct values for requested queries in input example provided in Figures~\ref{fig:amazon_example},~\ref{fig:input_example}.}
    \label{fig:output_example}
\end{figure}

\begin{figure*}[t]
    \centering
    \includegraphics[width=1.0\textwidth]{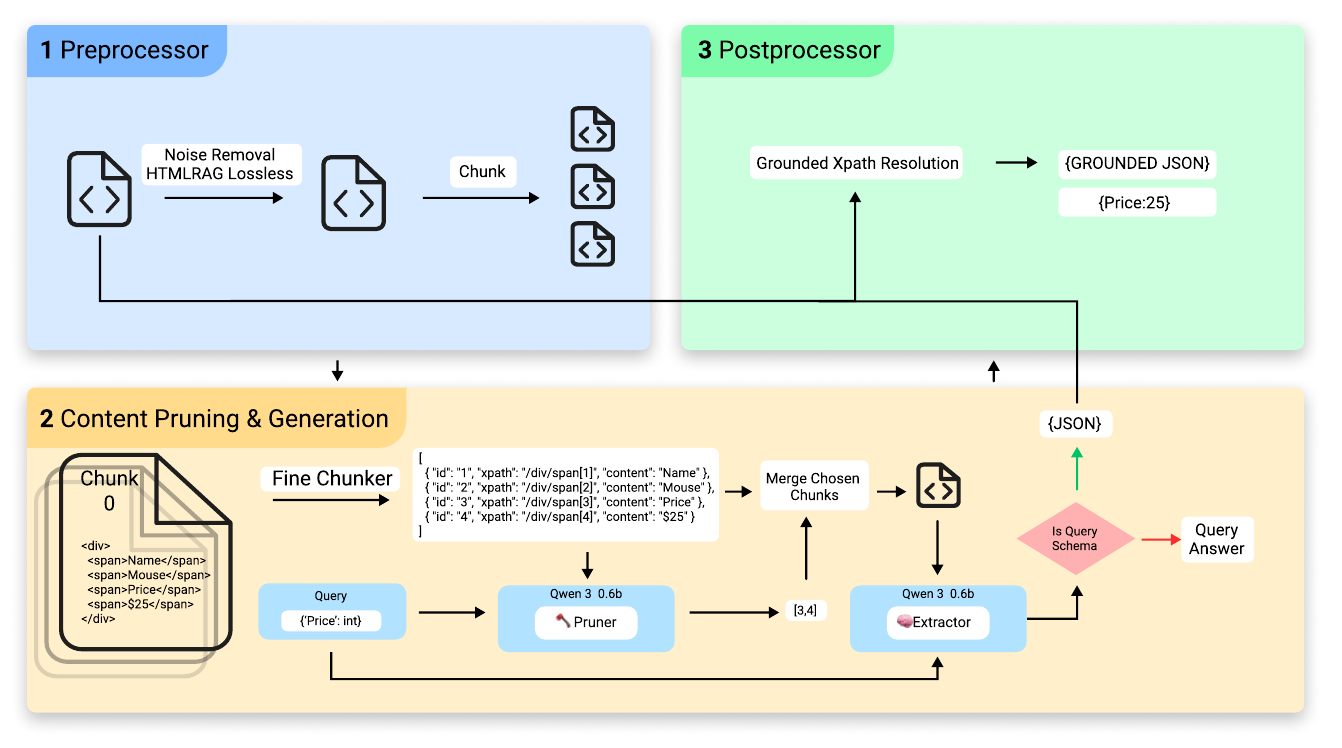}
    \caption{The AXE Pipeline Architecture: showing the flow from Raw HTML through AI-driven Pruning and Extraction to Grounded XPath Resolution.}
    \label{fig:overview}
\end{figure*}

Web pages usually have what is commonly known as \textbf{boilerplate} \cite{boilerplate} which is code added for functionality, aesthetics, or the business side of the website, not for content. Boilerplate includes components like the navigation bars, ads, footers, sections used for aesthetics, etc. The more boilerplate you can remove from the page, and the less irrelevant content the page has relative to your query, the easier it will be for a smaller LLM to perform well since it's fed less context.

We introduce \textbf{AXE} (Adaptive X-Path Extractor). \textbf{AXE} can be considered as a filter that removes boilerplate and parts of the HTML page that are irrelevant to your queries, to produce query-relevant, content-rich text that smaller LLMs can efficiently use to generate high-quality structured JSON output. \textbf{AXE} utilizes HTMLRAG \cite{HtmlRAG} to perform initial preprocessing then performs intelligent pruning on HTML nodes to prune the DOM Tree; finally, the distilled HTML page is fed into an LLM for generation.

Section ~\ref{Section2} of this paper provides a review of related works, telling the journey of methods evolving from heuristics to modern systems. Section ~\ref{sec:Methodology} details the \textbf{AXE} methodology and system architecture. Section ~\ref{sec:Results} reports our results. Section ~\ref{sec:ethics} mentions ethics regarding the usage of our system and data collection, while Section ~\ref{sec:conclusion} is the conclusion of our work. Section ~\ref{sec:Limitations} declares the limitations of our method.

The contributions of this paper are summarized as follows:
\begin{itemize}
    \item We propose \textbf{AXE}: A Low-Cost Cross-Domain Web Structured Information Extraction technique.
    \item We deliver 3 open-weight adaptors for Qwen3 0.6B LLM that vastly improve performance of the base model on our problem. 
    \item \textbf{AXE} achieves state-of-the-art zero-shot token-level F1 on SWDE dataset.
\end{itemize}

\section{Related Works}
\label{Section2}

Web Information Extraction (WIE) has evolved from structural heuristics to semantic, LLM-driven approaches. This section reviews this progression, categorizing existing methods into heuristic-based, machine learning, and LLM-integrated frameworks.

\subsection{Boilerplate Removal and Heuristics}
Traditional methods for web content extraction focus on general boilerplate removal for improved readability or summarization. Tools like Trafilatura \cite{Trafilatura} use rule-based cascades and XPath \cite{xpath} expressions, while Inscriptis \cite{Inscriptis} prioritizes layout preservation during HTML-to-text conversion. Other approaches treat extraction as a sequence labeling problem, using CNNs \cite{web2text} or hybrid systems like Dripper \cite{Dripper}, which combines heuristics with small LLMs for classification. Unlike these works, \textbf{AXE} targets specific query-relevant attributes rather than general content distillation, enabling more efficient pruning if you know what you're searching for.

\subsection{Structural and Pre-trained Models}
Recent models leverage HTML-specific structures to improve understanding. MarkupLM \cite{MarkupLM} incorporates XPath information into pre-training but requires domain-specific fine-tuning. WebLM \cite{WebLM} uses multi-modal visual rendering through Regions of Structure pooling to model DOM relationships. However, these specialized encoders often limit cross-domain adaptability—a core strength of AXE's architecture.

\subsection{LLM-based Retrieval}
The emergence of LLMs has shifted the focus toward RAG-based approaches. HTMLRAG \cite{HtmlRAG} demonstrates the superiority of web-aware context over plain text by leveraging DOM structure. While AXE shares this structural philosophy, it addresses HTMLRAG's computational bottlenecks by using aggressive pruning instead of additive XPath selection, specifically optimizing for structured data extraction over diverse schemas.

\begin{table*}[h]
  \centering
  \caption{Performance Comparison on WebSRC Dataset, *all methods are using Large variants of their methods if many variants of same method exist.}
  \begin{tabularx}{\textwidth}{X|l|cc|cc}
    \toprule
    \multirow{2}{*}{Method*} & \multirow{2}{*}{Modalities} & \multicolumn{2}{c|}{Dev} & \multicolumn{2}{c}{Test} \\
    \cmidrule(lr){3-4} \cmidrule(lr){5-6}
    & & EM$\uparrow$ & F1$\uparrow$ & EM$\uparrow$ & F1$\uparrow$ \\
    \midrule
    T-PLM(Electra) & Text                 & 61.67 & 69.85 & 56.32 & 72.35 \\
    H-PLM(Electra) & Text + HTML          & 70.12 & 74.14 & 66.29 & 72.71 \\
    V-PLM(Electra) & Text + HTML + Image  & 73.22 & 76.16 & 68.07 & 75.25 \\
    \hdashline
    LayoutLMv3  & Text + Image        & 71.38 & 75.73 & 57.68 & 63.33 \\
    MarkupLM    & Text + HTML         & 73.38 & 79.83 & 69.09 & 76.45 \\
    WebLM       & Text + HTML + Image & 78.40 & 84.24 & \textbf{72.01} & \textbf{78.66} \\
    \hdashline 
    \textbf{AXE (Ours)} & Text + HTML & \textbf{80.06} & \textbf{86.95} & 67.6 & 71.1 \\
    \bottomrule
  \end{tabularx}
  \label{tab:websrc}
\end{table*}

\section{Methodology: AXE}
\label{sec:Methodology}

\subsection{Overview}
AXE filters HTML content based on user-defined queries or schemas. It preprocesses raw HTML into chunks and uses a Pruning Adaptor to isolate relevant nodes. These are processed by QA or Schema Extraction Adaptors to produce structured results, which are then grounded to the original DOM tree to ensure structural fidelity.

\subsection{System Architecture}

AXE is divided into 3 main parts: preprocessor, AI extractor, and postprocessor. The preprocessor performs basic cleaning and chunking for the HTML text. The AI extractor is divided into two stages: one for pruning the HTML further and the other for the final generation output to produce the schema structure or the answer based on the question given. Finally, the postprocessor (in the case of schema extraction) checks whether the extracted content is actually grounded in HTML and not hallucinated; if so, it retrieves the text that best matches the extracted output.

\subsubsection{Preprocessor}
In the preprocessor, we first remove scripts and CSS from the page as they only add noise for text extraction. Then we use HTMLRAG lossless cleaning to further clean the HTML from unnecessary nested tags that consume extra tokens. From there, we use a specialized HTML chunking utility AutoChunker \cite{autochunker} to chunk the HTML into blocks instead of chunking it text-wise like normal text.

\subsubsection{AI Extractor}
After taking the cleaned list of chunks from the preprocessor, we further chunk each chunk into more refined mini-chunks, where each mini-chunk is a full XPath that ends with an atomic structure like a table list or a text block. That way, we have a list of XPaths for each chunk's content. 

\paragraph{Pruner}
From there, we pass it to the pruner to choose from the list of XPaths the relevant XPaths to the query. After extracting the relevant XPaths from all chunks in the HTML page, we re-merge the list of XPaths into a full HTML page to try to preserve the original flow of text. Then we rerun the HTMLRAG lossless compression to further remove any redundant tags in case we extract from a deep chunk. 

\paragraph{Extractor}
Finally, the merged HTML text and the query are fed to the LLM, activating either the QA Adaptor for question-style queries or the Schema Adaptor for schema/JSON-style queries. 

\subsubsection{Postprocessor}
After generating the LLM response, we perform additional post-processing to improve the system's recall. For the QA task, the answer is simply extracted and returned. However, for the schema, we run a grounding algorithm to ensure the answer is physically present in the source.

\paragraph{Grounded XPath Resolution (GXR)}
To mitigate the risk of "hallucinations" common in small language models, we implement a post-processing grounding mechanism termed Grounded XPath Resolution. While the model generates a textual answer, our system does not treat this as the final output. Instead, GXR performs a structural search across the source DOM tree to identify the node that most likely contains the predicted information.

The algorithm decomposes every HTML element into granular text chunks and employs a dual-objective scoring function: (1) Lexical Overlap, calculating the token intersection between the prediction and the candidate chunk; and (2) Fuzzy Semantic Similarity, using the Gestalt Pattern Matching ratio to account for minor character-level discrepancies. Once the optimal node is identified, the system resolves its unique absolute XPath. This process effectively transforms a generative task into a grounded retrieval task, ensuring that all extracted data is traceable to its original structural coordinate in the document.


\subsection{Training}
To produce three adaptors that are capable of the following tasks: Pruning, QA, and Schema extraction, we use knowledge distillation training. We used Qwen3-Coder-480B-A35B-Instruct as a teacher model to create synthetic schemas and questions with their respective answers to train the extraction adaptors, and the same teacher model to create a dataset for the model to learn how to prune.

\subsubsection{Data}
To get the diversity of data that is needed, we used Common Crawl to get a random set of 10k pages. Then we extracted different heuristics from each page. Using these heuristics, we performed deduplication by clustering those pages into 1,000 pages. From there, we further removed inappropriate pages, leading us to have 914. (Further details in the Appendix)

\paragraph{Schema Data}
To generate high-quality JSON extraction data, we use a temperature hyperparameter of 0.3 to ensure that the answer is deterministic to a certain degree while allowing for some creativity in the schemas generated. To ensure that the generated gold truth is actually correct, we verify that each extracted answer is present in the HTML page; if not, we discard the entire example.

\paragraph{QA Data}
For QA, things were a bit different. Since the original base model didn’t perform well in QA, we had to push it further. For that reason, while creating the synthetic dataset, we ensured to include the reasoning behind each answer to improve the reasoning capabilities of the model for this task.

\subsubsection{Finetuning}
We fine-tuned Low-Rank Adaptation (LoRA) \cite{lora} adaptors with a rank of 64 ($\approx$40M parameters) on the Qwen-0.6B backbone using Rank-Stabilized LoRA (rsLoRA) \cite{rslora}.

The \textbf{Pruner} adaptor was trained for 2 epochs with a 3,072 token context window, using NEFTune \cite{neftune} ($\alpha_{noise}=12$) and 0.2 dropout for robustness. We used a cosine scheduler with a peak learning rate of $1.7 \times 10^{-5}$ and a 10\% warmup.

The \textbf{Extractor} shared this backbone but used a 4,096 token context window and reduced regularization ($\alpha_{noise}=5$, 0.1 dropout) to prioritize high-fidelity JSON generation.

For the \textbf{QA-Specialized} task, we initialized from the Extractor weights and fine-tuned on 15k WebSRC samples for 3 epochs. To optimize performance, we utilized FlashAttention-2, Liger kernels, and pure BF16 precision, applying higher dropout (0.3) and weight decay (0.2) to improve generalization on multi-modal queries.

\section{Experiments Results}
\label{sec:Results}

In this section, we present a comprehensive evaluation of AXE, demonstrating its efficiency and effectiveness in web information extraction. We focus on three key aspects: (1) \textbf{Zero-shot generalization} across diverse domains, (2) \textbf{Cost efficiency} through massive token reduction, and (3) \textbf{Component contribution} via detailed ablation studies.

\subsection{Datasets \& Setup}
We evaluate AXE on two primary benchmarks: \textbf{SWDE} \cite{swde} and \textbf{WebSRC} \cite{websrc}. 

\textbf{SWDE} (Structured Web Data Extraction) comprises 124,291 pages across 80 websites and eight verticals (e.g., Auto, Book, Camera). It is the standard benchmark for evaluating cross-site generalization. We use the standard page-level F1 score metric. A key challenge in SWDE is the diversity of layouts; consistent performance across all verticals indicates robust structural understanding.

\textbf{WebSRC} (Web-based Structural Reading Comprehension) is a more complex, multi-modal benchmark with 400k+ question-answer pairs derived from 6,400 web segments. Unlike SWDE, which focuses on attribute extraction, WebSRC requires understanding the interplay between text, layout, and DOM hierarchy to answer natural language queries. We report the official F1 score and Exact Match (EM) metrics.

\subsection{Performance on SWDE}
Table~\ref{tab:swde} compares AXE's performance against state-of-the-art baselines. We evaluate AXE in a \textbf{zero-shot setting ($k=0$)}, meaning the model receives no site-specific training examples or seed websites. In contrast, most baselines, including MarkupLM and WebLM, rely on one-shot training ($k=1$) or more where they are exposed to a seed site or more from the same vertical to learn the extraction schema.

\textbf{AXE achieves a state-of-the-art F1 score of 88.10\%}, outperforming even the best supervised baseline, WebLM$_{\text{LARGE}}$ ($k=1$), which scores 87.57\%. This result highlights AXE's superior generalization capability. By "pruning" the DOM to its semantic core rather than overfitting to specific layout patterns, AXE eliminates the need for site-specific fine-tuning.

We also provide a breakdown of performance across SWDE verticals in Table~\ref{tab:individual_domains}. The system achieves consistent high performance, ranging from 80.90\% in the 'Auto' domain to 93.13\% in 'Restaurant', confirming its robustness across widely varying structural templates.

\begin{table}[H]
  \centering
    \begin{tabularx}{\columnwidth}{Xcc}
    \toprule
      Model \textbackslash{} \#Seed Sites & $k=0$$\uparrow$  & $k=1$$\uparrow$ \\ \midrule
      SSM                       & -                 & 63.00           \\
      Render-Full               & -                 & 84.30           \\
      FreeDOM-NL                & -                 & 72.52           \\
      FreeDOM-Full              & -                 & 82.32           \\
      SimpDOM                   & -                 & 83.06           \\
      \midrule                
      MarkupLM$_{\text{BASE}}$  & -                 & 82.11           \\
      MarkupLM$_{\text{LARGE}}$ & -                 & 85.71           \\
      WebLM$_{\text{BASE}}$     & -                 & 84.21           \\
      WebLM$_{\text{LARGE}}$    & -                 & \textbf{87.57}  \\
      \hdashline
      \textbf{AXE (ours)}       & \textbf{88.10}   & -               \\
      \bottomrule
    \end{tabularx}
  \caption{F1 score on SWDE (all domains). Our system (zero-shot, $k=0$) outperforms the strongest one-shot ($k=1$) baselines, demonstrating superior cross-domain generalization without requiring seed examples.}
  \label{tab:swde}
\end{table}

\begin{table}[H]
    \centering
    \begin{tabularx}{\columnwidth}{Xc}
        \toprule
        Domain & F1 (\%) \\
        \midrule
        Auto & 80.90 \\
        University & 84.11 \\
        Camera & 87.00 \\
        Book & 92.42 \\
        Job & 88.96 \\
        NBA Player & 89.35 \\
        Movie & 88.93 \\
        Restaurant & 93.13 \\
        \midrule
        \textbf{Average} & \textbf{88.10} \\
        \bottomrule
    \end{tabularx}
    \caption{Performance of AXE across different SWDE domains, showing consistent effectiveness across varying website structures.}
    \label{tab:individual_domains}
\end{table}

\subsection{Efficiency Analysis: The Power of Pruning}
The core hypothesis of AXE is that most HTML content is noise. Table~\ref{tab:token_reduction} quantifies the efficiency of our Pruning Adaptor. On the SWDE dataset, the pruner reduces the average context length from 16,581 tokens to just 350.6 tokens—a \textbf{97.9\% reduction}. 

This massive reduction is crucial for two reasons: (1) It allows the use of a lightweight 0.6B parameter model, as the relevant context is dense enough to be processed purely; and (2) it drastically reduces inference cost and latency. As shown in the ablation study (Table~\ref{tab:ablation_swde}), removing the pruner ("w/o Pruner") results in a negligible drop in accuracy (from 88.37\% to 87.71\%), proving that the pruning is highly distinct and preserves virtually all task-relevant information.

\begin{table}[H]
    \centering
    \begin{tabularx}{\columnwidth}{Xc} 
        \toprule
        Method & \#Tokens \\
        \midrule
        Preprocessed    & 16581.9 \\
        After Pruning   & 350.6 ($\downarrow$97.9\%) \\
        \bottomrule
    \end{tabularx}
    \caption{Reduction of tokens by the pruner on SWDE dataset. The pruning step eliminates nearly 98\% of the context while retaining essential information.}
    \label{tab:token_reduction}
\end{table}

\subsection{Ablation Studies}
We conducted extensive ablation studies on both SWDE and WebSRC to validate the contribution of each module.

\subsubsection{Impact of Components on SWDE}
Table~\ref{tab:ablation_swde} breaks down the contribution of AXE's components.
\begin{enumerate}
    \item \textbf{Pruner:} As noted, removing the pruner ("w/o Pruner") causes a slight 0.66\% drop, confirming that while pruning is essential for efficiency, the model is robust enough to handle the noise if necessary (at the cost of much higher compute).
    \item \textbf{Grounded XPath Resolution (GXR):} Removing the GXR algorithm ("w/o Exact Algorithm") results in a significant 4.42\% drop (88.37\% $\rightarrow$ 83.95\%). This underscores the importance of our grounding mechanism; generating text alone is insufficient for precise structural extraction. The model must be forced to point to the exact source node.
    \item \textbf{Adaptation:} "w/o Adaptor" shows a 4.86\% drop, highlighting the effectiveness of our specialized LoRA fine-tuning.
\end{enumerate}

\begin{table}[H]
    \centering
    \begin{tabularx}{\columnwidth}{Xc}
        \toprule
        Method & F1 (\%) \\
        \midrule
        \midrule
        $\text{AXE (Full)}$ & 88.37 * \\
        \quad - w/o Pruner & 87.71 ($\downarrow$0.66\%) \\
        \quad - w/o Exact Algorithm & 83.95 ($\downarrow$4.42\%) \\
        \quad - w/o Adaptor & 83.51 ($\downarrow$4.86\%) \\
        \quad - w/o AXE Extraction& 78.17($\downarrow$10.2\%)  \\
        \bottomrule
    \end{tabularx}
    \caption{Ablation study on SWDE. The significant drops when removing GXR ("Exact Algorithm") and Finetuning ("Adaptor") validate our architectural choices. *Note that this result is based on 1000 samples from each domain, unlike Table ~\ref{tab:individual_domains} and Table~\ref{tab:swde} which have results with all of the dataset for fair comparison.}
    \label{tab:ablation_swde}
\end{table}

\subsubsection{Impact of Components on WebSRC}
For WebSRC (Table~\ref{tab:ablation_websrc}), which requires complex reasoning, the results are similar. Removing the QA Finetuning drops performance by over 11\% (86.95\% $\rightarrow$ 75.43\%), indicating that the base model requires specific training to align with the Question-Answering format of structural data. Removing Extraction Finetuning causes a catastrophic drop to 62.67\%, confirming that the specialized extraction capabilities are fundamental to the system's performance.

\begin{table}[H]
    \centering
    \begin{tabularx}{\columnwidth}{Xc} 
        \toprule
        Method & F1 (\%) \\
        \midrule
        $\text{AXE (Full)}$ & 86.95 \\
        \quad -w/o QA Finetuning & 75.43 ($\downarrow$11.52\%) \\
        \quad -w/o Extraction Finetuning & 62.67 ($\downarrow$24.28\%) \\
        \bottomrule
    \end{tabularx}
    \caption{Ablation study on WebSRC (Dev set). Specialized finetuning is critical for the complex QA tasks in WebSRC.}
    \label{tab:ablation_websrc}
\end{table}

\subsection{Impact of Chunk Size}
Finally, we analyzed the impact of the HTML chunk size on performance (Figure~\ref{fig:chunk_size_analysis}). Performance increases with chunk size, as larger chunks provide more local context (e.g., surrounding siblings or parent labels) that helps disambiguate similar nodes. The performance plateaus around 3000-4000 tokens, suggesting this is the optimal sweet spot between context sufficiency and processing granularity.

\definecolor{darkblue}{RGB}{74, 89, 111}
\definecolor{lightblue}{RGB}{155, 177, 199}
\begin{figure}[H]
    \centering
    \begin{tikzpicture}
        \begin{axis}[
            ybar,
            width=\columnwidth,
            height=7cm,
            bar width=18pt,
            axis x line*=bottom,
            axis y line*=left,
            ymajorgrids=true,
            grid style={line width=.1pt, draw=gray!20},
            xlabel={\small \textbf{Chunk Size}},
            ylabel={\small \textbf{F1 Score}},
            symbolic x coords={500, 1000, 2000, 3000, 4000, 5000},
            xtick=data,
            yticklabel style={/pgf/number format/fixed, /pgf/number format/precision=2},
            ymin=0.84,
            ymax=0.90,
            enlarge x limits=0.15,
            cycle list={
                {fill=darkblue, draw=none},
            },
            nodes near coords,
            every node near coord/.append style={
                font=\scriptsize\sffamily,
                inner sep=2pt,
                fill=white,          
                fill opacity=0.7,
                text opacity=1,
                /pgf/number format/.cd,
                fixed,
                precision=4
            },
        ]
            \addplot coordinates {
                (500, 0.85699)
                (1000, 0.86405)
                (2000, 0.87665)
                (3000, 0.88663)
                (4000, 0.88726)
                (5000, 0.88657)
            };
        \end{axis}
    \end{tikzpicture}
    \caption{Impact of chunk size on SWDE extraction performance. Larger chunks provide beneficial context, with performance peaking around 4000 tokens.}
    \label{fig:chunk_size_analysis}
\end{figure}

\section{Ethics}
\label{sec:ethics}
The primary dataset for training was sourced from Common Crawl, a public web archive, and we followed standard procedures for deduplication and filtering to ensure data quality and safety. While web-crawled data may contain sensitive information, our system is designed for structural extraction and does not store or process personally identifiable information (PII) beyond what is publicly accessible. We encourage users of our methodology to respect website \texttt{robots.txt} files and adhere to copyright and privacy regulations when deploying these tools.

\section{Conclusion}
\label{sec:conclusion}
In this paper, we presented AXE, demonstrating that effective web data extraction can be achieved without excessive computational resources or increasingly large models. By prioritizing the structural representation of information and systematically pruning irrelevant boilerplate content, we transform complex generation tasks into streamlined, grounded workflows. Our evaluations on the SWDE and WebSRC datasets indicate that a concise, well-structured DOM representation is often more beneficial for model performance than simply utilizing larger context windows. While AXE can be further enhanced by incorporating visual rendering features, it represents a significant advancement toward efficient and practical web mining. We provide our open-weight adaptors and the GXR methodology as a contribution to the community, facilitating structured information extraction from complex web environments.

\section{Limitations}
\label{sec:Limitations}
While effective, \textbf{AXE} primarily extracts textual information and does not make use of visual rendering. It also currently only supports HTML documents. Furthermore, our reliance on synthetic datasets for training may introduce domain-specific biases. Finally, while we utilize GXR to improve recall, the model is still susceptible to potential hallucinations; if the retrieved context is of poor quality, the extraction process will likely fail.






\bibliography{custom}

\section*{Appendix}

\begin{algorithm*}
\caption{Find Closest HTML Node with Fuzzy Matching}
\begin{algorithmic}[1]
\Require $H$: The HTML source text
\Require $S$: The search text string
\Ensure A tuple containing the best matching chunk text, XPath, sub-index, and score.

\Function{FindClosestNode}{$H, S$}
    \State $S_{norm} \gets \textsc{Normalize}(S)$
    \If{$S_{norm}$ is empty}
        \State \Return $\{ \text{found}: \textbf{False}, \text{score}: 0 \}$
    \EndIf

    \State $D \gets \textsc{ParseDOM}(H)$ \Comment{Initialize BeautifulSoup}
    \State $score_{best} \gets 0$
    \State $subset_{best} \gets 0$
    \State $match_{best} \gets \textbf{None}$

    \ForAll{element $E \in D$}
        \State $C \gets \textsc{GetTextChunks}(E)$ \Comment{Split element content by tags}
        \For{$i \gets 0$ \textbf{to} length of $C$}
            \State $c \gets C[i]$
            \State $c_{norm} \gets \textsc{Normalize}(c)$
            \State $tokens_{S} \gets \textsc{Split}(S_{norm})$
            \State $tokens_{c} \gets \textsc{Split}(c_{norm})$
            \State $I \gets |tokens_{S} \cap tokens_{c}|$ \Comment{Count intersection tokens}

            \If{$S_{norm} \subseteq c_{norm} \lor I > 0$}
                \State $sim \gets \textsc{SequenceMatcher}(c, S)$ \Comment{Compute Fuzzy Score}
                
                \If{$sim \geq score_{best} \land I \geq subset_{best}$}
                    \State $score_{best} \gets sim$
                    \State $subset_{best} \gets I$
                    \State $match_{best} \gets \{ \text{elem}: E, \text{text}: c, \text{idx}: i \}$
                \EndIf
            \EndIf
        \EndFor
    \EndFor

    \If{$match_{best}$ is \textbf{None}}
        \State \Return $\{ \text{found}: \textbf{False}, \text{score}: 0 \}$
    \Else
        \State $path \gets \textsc{GetXPath}(match_{best}.\text{elem})$
        \State \Return $\{ \text{text}: match_{best}.\text{text}, \text{xpath}: path, \text{sub\_index}: match_{best}.\text{idx}, \text{score}: score_{best} \}$
    \EndIf
\EndFunction
\label{fig:algo}
\end{algorithmic}
\end{algorithm*}

\begin{figure*}[h]
    \centering
    \includegraphics[width=0.8\textwidth]{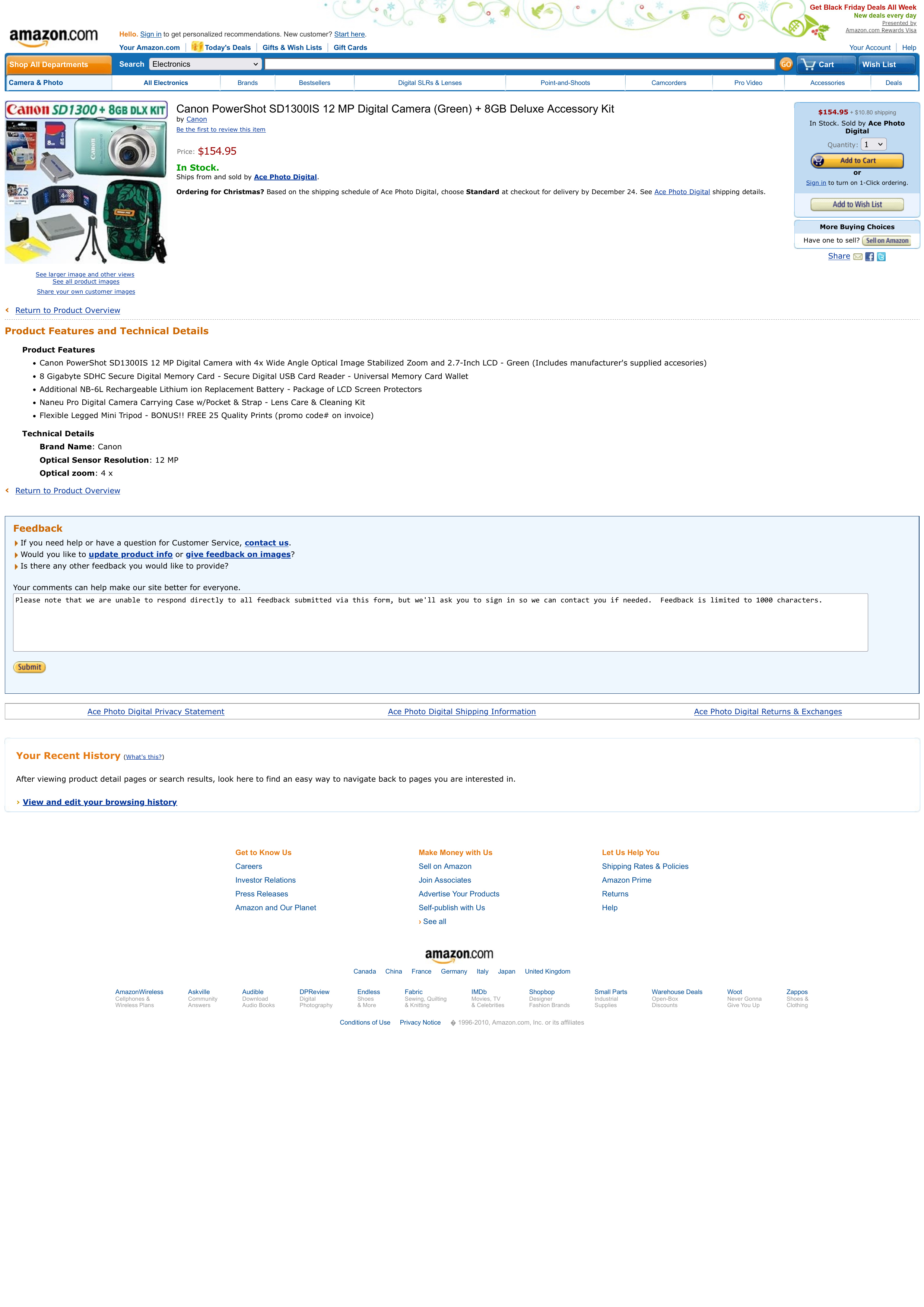}
    \caption{An Amazon web page example from SWDE Dataset before applying the pruner}
    \label{fig:before_pruner}
\end{figure*}


\begin{figure*}[h]
    \centering
    \begin{codebox}{Input Schema}
    \begin{lstlisting}[language=json]
$154.95 + $10.80 shipping
Canon PowerShot SD1300IS 12 MP Digital Camera
(Green) + 8GB Deluxe Accessory Kit by Canon
$154.95
    \end{lstlisting}
    \end{codebox}
    \caption{Same Amazon web page example from SWDE Dataset from Figure~\ref{fig:before_pruner} after applying the pruner}
    \label{fig:after_pruner}
\end{figure*}

\begin{table}[h]
\centering
\caption{Hyperparameters for AXE Pruner Adaptor Training}
\label{tab:hypers}
\begin{tabular}{ll}
\toprule
\textbf{Hyperparameter} & \textbf{Value} \\
\midrule
Base Model & Qwen-0.6B \\
Fine-tuning Method & LoRA (rsLoRA) \\
LoRA Rank ($r$) & 64 \\
LoRA Alpha ($\alpha$) & 64 \\
LoRA Target Modules & All Linear Modules \\
LoRA Dropout & 0.2 \\
NEFTune Alpha & 12 \\
Max Sequence Length & 3,072 tokens \\
\midrule
Optimizer & AdamW \\
Learning Rate & $1.7 \times 10^{-5}$ \\
LR Scheduler & Cosine \\
Warmup Ratio & 0.1 \\
Effective Batch Size & 16 \\
Training Epochs & 2.0 \\
Weight Decay & 0.1 \\
Precision & FP16 \\
\bottomrule
\end{tabular}
\end{table}

\begin{table}[ht]
\centering
\caption{Hyperparameters for the AXE QA-Specialized Adaptor Fine-tuning.}
\label{tab:qa_hypers}
\begin{tabular}{ll}
\toprule
\textbf{Hyperparameter} & \textbf{Value} \\
\midrule
Base Model & Qwen-0.6B \\
Fine-tuning Method & rsLoRA \\
LoRA Rank ($r$) & 64 \\
LoRA Alpha ($\alpha$) & 64 \\
LoRA Target Modules & All Linear Modules \\
LoRA Dropout & 0.3 \\
Max Context Length & 4,096 tokens \\
\midrule
Optimizer & AdamW \\
Learning Rate & $1 \times 10^{-5}$ \\
LR Scheduler & Cosine \\
Warmup Ratio & 0.1 \\
Weight Decay & 0.2 \\
Effective Batch Size & 32 \\
Training Epochs & 2.0 \\
\midrule
Precision & Pure BF16 \\
Attention Mechanism & FlashAttention-2 \\
Kernel Optimization & Liger Kernel \\
NEFTune Alpha & 8 \\
\bottomrule
\end{tabular}
\end{table}

\begin{table*}[ht]
\centering
\caption{Hyperparameters used for fine-tuning the Extractor and Adaptor modules.}
\label{tab:hyperparams}
\begin{tabularx}{\textwidth}{Xcc} 
\toprule
\textbf{Category} & \textbf{Parameter} & \textbf{Value} \\
\midrule
\multirow{3}{*}{\textbf{Model}} 
 & Base Model & Qwen/Qwen3-0.6B \\
 & Context Length & 4096 \\
 & Precision & FP16 \\
\midrule
\multirow{6}{*}{\textbf{LoRA Method}} 
 & Rank ($r$) & 64 \\
 & Alpha ($\alpha$) & 64 \\
 & Dropout & 0.1 \\
 & Target Modules & All \\
 & Use RSLoRA & True \\
 & NEFTune Noise $\alpha$ & 5 \\
\midrule
\multirow{9}{*}{\textbf{Training}} 
 & Learning Rate & $1.7 \times 10^{-5}$ \\
 & Weight Decay & 0.1 \\
 & Batch Size (Per Device) & 2 \\
 & Gradient Accumulation & 8 \\
 & Effective Batch Size & $16 \times N_{GPU}$ \\
 & Scheduler & Cosine \\
 & Warmup Ratio & 0.1 \\
 & Epochs & 2 \\
 & Kernel Optimization & Liger Kernel \\
\bottomrule
\end{tabularx}
\end{table*}

\begin{figure*}[h]
    \centering
    \begin{codebox}{Pruner Prompt}
    \begin{lstlisting}[language=json]
You are a Smart and Clever Context Selector. Your task is to filter a list of HTML chunks, keeping ONLY the ones relevant to the provided Query/Schema and any necessary context to answer the query.

Query/Schema:
{query}

**INSTRUCTIONS:**

1.  **Analyze the Query:** Determine exactly what data is being requested. It could be specific content (prices, dates), structural elements (menu items, footers), or broad sections.
2.  **Select Relevant Chunks:** Identify chunks that contain:
    *   The **Direct Answer** (values, text, list items).
    *   Essential **Labels/Context** (e.g., the text "Price:" next to "$10.00").
    *   **Atomic Containers** (tables, lists) that hold the requested data.
3. **Select Context Carefully:** Only include chunks that are necessary to understand or locate the answer. Avoid including unrelated sections.
4.  **Discard Noise:** Remove any chunks that do not contribute to answering *this specific query*.
5.  **Handle Missing Data:** If no chunks contain the requested information, return an empty list `[]`.
6. **Include Supporting Context:** When relevant, include chunks that provide necessary context to understand or locate the answer, even if they don't contain the direct answer themselves.
7. **Table Handling:** If the query relates to tabular data, prioritize chunks that represent entire rows or columns relevant to the schema.
8. **Flow**: Ensure the selected chunks form a coherent context for answering the query.

**Content:**
{content}

**Response Format:**
Output ONLY a valid JSON list of indices.
Example: [1, 4, 12] or []
    \end{lstlisting}
    \end{codebox}
    \caption{Pruner Prompt}
    \label{fig:prompts:pruner}
\end{figure*}

\begin{figure*}[h]
    \centering
    \begin{codebox}{Extractor Prompt}
    \begin{lstlisting}[language=json]
You are an expert Data Extraction and ETL agent. Your task is to parse the provided HTML content and extract specific data points to populate a target JSON schema.

Target Schema Structure:
{query}

HTML Content:
{content}

RULES:
1. Extract exact substrings from the text content of the HTML. Do not invent data.
2. Ignore HTML tags, attributes, and styles; extract only the visible text value.
3. If a specific field from the schema is not found in the content, set its value to null.
4. Ensure the output strictly follows the keys defined in the "Target Schema Structure".
5. Your output MUST be exactly as shown in the HTML.
6. Be concise and avoid adding any extra information outside the schema.

OUTPUT FORMAT:
REASONING: "The reasoning behind the answer"
{{json filled schema"}}
    \end{lstlisting}
    \end{codebox}
    \caption{Extractor Prompt}
    \label{fig:prompts:extractor}
\end{figure*}

\begin{figure*}[h]
    \centering
    \begin{codebox}{QA Extractor Prompt}
    \begin{lstlisting}[language=json]
You are a highly precise Context-Aware Question Answering engine. Your sole task is to extract the answer to the User Query based ONLY on the provided Context.

User Query:
{query}

Context:
{content}

INSTRUCTIONS:
1. Answer the query using ONLY information found in the Context. Do not use outside knowledge.
2. If the answer is not present in the Context, set the value to null.
3. Your output must be valid, parseable JSON.
4. Provide concise answers without additional commentary.
5. If the query is boolean, respond with yes or no.
6. Choose the most relevant information if multiple answers exist.

OUTPUT FORMAT:
REASONING: "The reasoning behind the answer"
{{"answer": "The extracted text or synthesized answer"}}
    \end{lstlisting}
    \end{codebox}
    \caption{QA Extractor Prompt}
    \label{fig:prompts:QA}
\end{figure*}

\end{document}